\documentclass[letterpaper, 10pt, conference]{ieeeconf}

\IEEEoverridecommandlockouts                  

\usepackage{tikz}
\usepackage{siunitx}
\usepackage{xspace}
\usepackage{wrapfig}
\usepackage{graphicx}
\usepackage{booktabs}
\usepackage{amsmath}

\usepackage{ucs}
\usepackage[utf8x]{inputenc}
\usepackage{caption}
\usepackage{fontenc}
\usepackage{xcolor}
\usepackage{subfigure}

\usepackage[font=small,labelfont=bf]{caption}
\usepackage{hyperref}  
\hypersetup{
    colorlinks=true,
    linkcolor=blue,
    filecolor=magenta,      
    urlcolor=blue,
    citecolor=blue,
}
\newcommand{\figref}[1]{Fig.~\ref{#1}}

\newcommand{\secref}[1]{Section~\ref{#1}}

\newcommand{\hamrsimple}{\textit{HAMR-simple}\xspace}
\newcommand{\hamrfull}{\textit{HAMR-full}\xspace}

\DeclareMathOperator*{\argmin}{arg\,min}

\usetikzlibrary{arrows, positioning,fit,calc}

\usepackage{amssymb}

\title{\LARGE \bf
Residual Model Learning for Microrobot Control
}

\author{Joshua Gruenstein$^{1}$, Tao Chen$^{1}$, Neel Doshi$^{2}$, and Pulkit Agrawal$^{1}$%
\thanks{The authors are with $^{1}$ the Improbable AI Lab, which is part of the Computer Science and Artificial Intelligence Laboratory (CSAIL) at Massachusetts Institute of Technology and $^{2}$ the Department of Mechanical Engineering at Massachusetts Institute of Technology, Cambridge, MA 02139. 
\tt{\small{\{jgru, taochen, nddoshi, pulkitag\}@mit.edu}}}%
}

\begin{document}

\maketitle
\thispagestyle{empty}
\pagestyle{empty}

\begin{abstract}
A majority of microrobots are constructed using compliant materials that are difficult to model analytically, limiting the utility of traditional model-based controllers. Challenges in data collection on microrobots and large errors between simulated models and real robots make current model-based learning and sim-to-real transfer methods difficult to apply. We propose a novel framework \textit{residual model learning} (RML) that leverages approximate models to substantially reduce the sample complexity associated with learning an accurate robot model. We show that using RML, we can learn a model of the Harvard Ambulatory MicroRobot (HAMR) using just 12 seconds of passively collected interaction data. The learned model is accurate enough to be leveraged as ``proxy-simulator'' for learning walking and turning behaviors using model-free reinforcement learning algorithms. RML provides a general framework for learning from extremely small amounts of interaction data, and our experiments with HAMR clearly demonstrate that RML substantially outperforms existing techniques.
\end{abstract}
\section{Introduction}
Autonomous microrobots (i.e., millimeter-scale robots)  offer the promise of low-cost exploration and monitoring of places that are impossible for their larger, conventional counterparts (or humans) to reach. However, the benefits of small size come at a high cost. While traditional robots are built using rigid materials with well-understood manufacturing processes, most microrobots are built out of compliant materials using relatively novel manufacturing processes, including laminate manufacturing and 3D-printing. Though the use of novel manufacturing paradigms is necessary to generate reliable motion at the millimeter scale, modeling flexible materials is challenging. Thus, a large majority of current microrobots rely on either hand-crafted locomotion primitives~\cite{goldberg2017highb} or on controllers that limit the scope of possible robot behaviors~\cite{doshi2018contact}. 

\begin{figure}[t]
    \centering
    \includegraphics{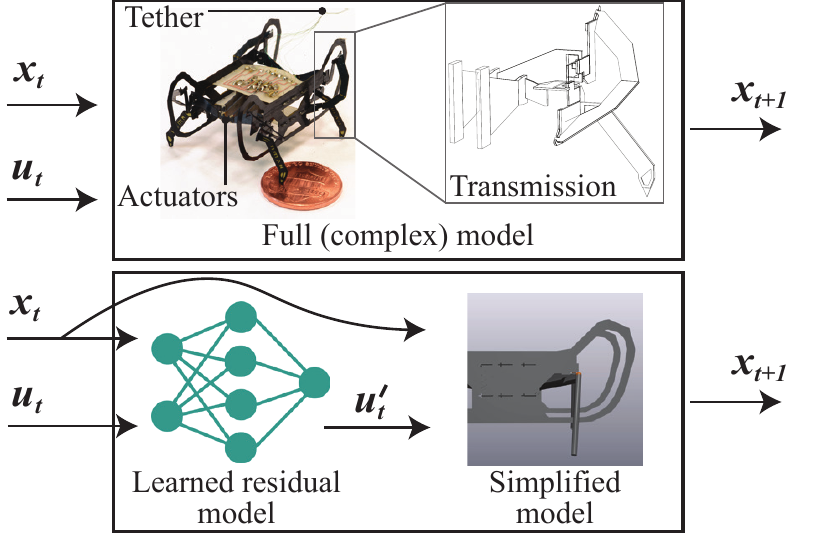}
    \caption{An overview of our \textit{residual model learning} framework.  Rather than modeling the full robot dynamics all at once (\textbf{Top}), we decompose the dynamics into two distinct components (\textbf{Bottom}): a learned residual model providing fast execution of complex transmission dynamics, and a simplified analytical model covers contacts and rigid body dynamics. The result is a 47$\times$ speedup in model execution time, while retaining sufficient accuracy for zero-shot transfer of RL policies back to the original full model.}
    \vspace{-0.4cm}
    \label{fig:intro_fig}
\end{figure}

In recent years, model-free deep reinforcement learning methods have enjoyed tremendous success on various simulated tasks~\cite{Mnih15,Silver16,heess2016learning,schulman2017proximal,li19relationalrl}. However, because these methods require millions of data samples, they are infeasible for real-world application. 
To reduce the real-world data requirements, broadly speaking, two approaches have been pursued: (a) model-based reinforcement learning (MBRL~\cite{nagabandi2018neural,chua2018deep}) and (b) sim-to-real transfer~\cite{tan2018sim}. 

While MBRL is more sample efficient than model-free algorithms, it still requires few hours of real-world data collection~\cite{nagabandi2018neural}, which is often infeasible for fragile microrobots. Therefore, to improve sample efficacy, prior MBRL approaches for microrobots learn in a restricted action space \cite{yang2018learning}, which inadvertently reduces robot agility and precludes learning of complex behaviors. Furthermore, sim-to-real transfer is challenging due to slow and/or inaccurate simulation. Most robots at the millimeter-scale are often designed using manufacturing methods (e.g., laminate manufacturing~ \cite{hoover2008fast, sreetharan2012monolithic}) that approximate traditional joints (e.g., pin-joints, universal-joints, etc.) with compliant flexures. These flexure-based transmissions are then difficult to model using traditional rigid or pseudo-rigid body approaches~\cite{howell2013compliant, odhner2012smooth}, resulting either in inaccurate approximate models \cite{finio2011system} or computationally expensive high-fidelity models~\cite{doshi2018contact, hoffman2011passive}. 

We propose a novel framework called \textit{residual model learning} (RML) that learns accurate models using a fraction of the data required by existing model-based learning techniques. Our key insight is as following: typical learning methods are end-to-end and aim to predict the robot’s next state from the current state and the applied actuation. This means that these models not only need to account for non-linear transmission (owing to robot’s construction with non-rigid materials), but also learn contact dynamics of the robot with the surface it walks on. Contacts generate external forces that affect the robot’s motion through the transmission, but are otherwise decoupled from the transmission dynamics. Because it is fast to simulate contacts with the rigid transmission, the focus of learning can solely be to model the residual between the rigid and non-rigid transmission and not on learning the contact dynamics or the rigid transmission. E.g., one can learn to predict residual forces that, when applied to a rigid-body simulator, would result in the same next state as if the model was non-rigid. Predicting these residual forces will require much less data than learning the coupled non-rigid transmission and the contact dynamics model end to end. 

The learned residual model can be thought of as performing system identification and can be used as a ``fast simulator" of the real robot. This learned simulator can be used to train model-free policies that optimize the task reward.  We experimentally test the proposed framework on a simulated version of the Harvard Ambulatory Microrobot (HAMR \cite{baisch2011design}, ~\figref{fig:intro_fig}, top) a challenging system due to its complex transmissions, flexure-based joints, and non-linear actuation. Due to the ongoing pandemic, we did not have access to the lab space to run real-world experiments. To reflect the real world as closely as possible, we consider two simulated versions of HAMR: \hamrsimple and \hamrfull. \hamrsimple is an approximate model and runs 200x times faster than \hamrfull, which emulates the actual robot in much greater detail. These models are described in Section~\ref{sec:setup}. One can consider \hamrsimple akin to the analytical simulation model and the more complex \hamrfull akin to the real robot. 

Using RML, we can learn a model of \hamrfull with only \SI{12}{\second} of passively collected data while the robot moves in free space (~\secref{sec:method}). This learned model runs 47x times faster than  \hamrfull and is accurate enough to be used as a ``proxy-simulator" for training model-free RL policies. Experiments in~\secref{sec:experiments} show that the learned \hamrfull model can be used to train locomotion policies that transfer to the true \hamrfull simulator for walking forward and backward as well as turning left and right on flat terrain.

In comparison to MPC techniques, our method allows for real-time control of microrobots as the learned model-free policy is inexpensive to evaluate. It is possible to run it at the required \SI{250}{\hertz} on embedded controllers. In comparison to prior methods, RML outperforms RL policies trained using \hamrsimple, state-of-the-art MBRL approach~\cite{chua2018deep} and various domain randomization techniques. Overall, our results demonstrate that RML significantly outperforms prior methods and is a data-efficient learning-based method for controlling microrobots.

\section{Related Work}

\textbf{Legged Locomotion}:
Controlling legged locomotion usually relies on trajectory optimization~\cite{posa2014direct, medeiros2020trajectory, carius2019trajectory, apgar2018fast}, which requires an accurate model of the robot's dynamics. However, it is generally difficult to build an accurate dynamic model of microrobot locomotion that is computationally tractable, i.e, results in an efficiently solvable trajectory optimization \cite{doshi2018contact}. An alternative approach is to use model-free methods, which often require good simulators or must be trained in the real world. Recent work in model-free legged locomotion learning in the real world \cite{ha2020learning} is not currently applicable to microrobots such as HAMR, as their fragile construction would limits sufficient exploration without risk of damage. Our approach offers a compromise between the two: we build a simplified robot model, learn a residual model to match the full complex robot model, and finally use model-free RL algorithms to optimize the control.

\textbf{Microrobots}: Prior work on controlling microrobot locomotion has generally relied on tuned passive dynamics  that can potentially limit the versatility of the robot \cite{bailey2001comparing} or the design of experimentally informed controllers that are time-consuming to tune \cite{goldberg2017highb}. More recently, drawing inspiration from the control of larger legged robots, trajectory optimization \cite{doshi2018contact} and policy learning methods \cite{yang2018learning} have also been applied. However, as mentioned above, prior models (such as \hamrfull) are costly to evaluate and intractable for compute-intensive learning methods such as model-free RL algorithms. Other work for microrobots has solved for lower-dimensional policy representations (e.g., gait patterns using Bayesian optimization \cite{yang2018learning}) that require fewer simulated samples. However, this significantly limits the robot's agility and disallows more complex behaviors such as jumping or angled walking, and leads to slower and less efficient policies. We build on these approaches, developing a computationally efficient high-fidelity model that is suitable for both trajectory optimization and model-free RL.

\textbf{Sim-to-Real}:  A common method to overcome data scarcity in RL is to train a learning system in simulation then transfer the learned policy to the real world. However, differences between the simulation and the real world (the \textit{sim-to-real gap}) impede transfer quality. Recent works narrow the sim-to-real gap by using domain randomization~\cite{tobin2017domain, james2017transferring, tremblay2018deep, peng2018sim, chen2018hardware}, performing system identification~\cite{kolev2015physically, yu2017preparing}, exploiting more transferable visual feature representation~\cite{mahler2017dex, muller2018driving, pomerleau1989alvinn}, improving the simulation fidelity by matching robot trajectories in simulation and in the real world~\cite{chebotar2019closing}, and mapping observations, states, and actions between simulation and the real world~\cite{hanna2017grounded, christiano2016transfer, golemo2018sim, bousmalis2018using}. Our work also partially learns a dynamics model to facilitate the transfer. However, our residual model only learns the compliant dynamics of the flexure-based transmission, relying on a rigid body model to resolve contacts, which makes the dynamics model learning much more data-efficient.

\textbf{Actuator learning}: Some previous work integrates a learned actuator model into an otherwise analytical rigid-body simulation, such as with the ANYmal quadruped~\cite{hwangbo2019learning}.  This approach is conceptually similar to our hybrid modeling approach but is much more limited in scope.  While Hwangbo et. al \cite{hwangbo2019learning}, characterize a single, difficult-to-model element (ANYmal's actuators) through system identification, our approach learns from complete robot data and can correct for a variety of modeling errors that accumulate from the misidentification of physical components.

\textbf{Residual Learning}: Learning a residual model has been explored in a few previous works. Several researchers use learning to compensate for the state prediction error of the analytical models~\cite{fazeli2017learning, ajay2019combining, fazeli2020long}. Another approach is to learn a residual policy that is added on top of a base policy, which can be either an analytical controller, a hand-engineered controller, or some model-predictive controller~\cite{zeng2020tossingbot, silver2018residual, johannink2019residual}. In contrast to these works, our approach learns a residual model that projects the actions in the action space of a complex dynamics model into the action space of a simplified dynamics model such that both models transit to the same future states.
\section{Platform Description and Preliminaries}
\label{sec:setup}

\begin{figure}[t]
    \centering
    \includegraphics{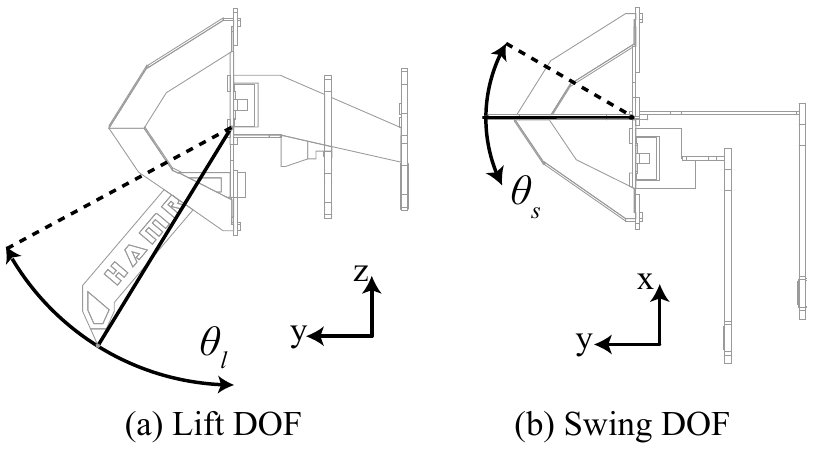}
    \caption{Two-dimensional projections of HAMR's SFB-transmission showing the (a) lift and (b) swing DOF in gray. The kinematics of \hamrsimple's transmission is overlayed in black.}
    \label{fig:trans_schematic}
    \vspace{-0.2cm}
\end{figure}

HAMR (\figref{fig:intro_fig}) is a \SI{4.5}{\centi\meter} long, \SI{1.43}{\gram} quadrupedal microrobot with eight independently actuated degrees-of-freedom (DOFs). Each leg has two DOFs that are driven by optimal energy density piezoelectric bending actuators \cite{jafferis2015design}. A spherical-five-bar (SFB) transmission connects the two actuators to a single leg in a nominally decoupled manner: the lift actuator controls the leg's vertical position (\figref{fig:trans_schematic}a), and the swing actuator controls the leg's fore-aft position (\figref{fig:trans_schematic}b). 

The dynamic model of HAMR previously used for trajectory planning and control~\cite{doshi2018contact} consists of four SFB-transmissions attached to the robot's (rigid) chassis. We call this model \hamrfull. In an effort to improve transfer to the physical robot, this model explicitly simulates the dynamics of the SFB-transmission. This, however, is computationally intensive due to the complex kinematics of the SFB-transmissions (\secref{sec:hamr-full}, \figref{fig:trans_schematic}).

A simpler lumped-parameter approximation of the SFB-transmissions can be constructed using the insight that its kinematics constrains the motion of each leg to lie on the surface of a sphere. Thus, we can treat each SFB transmission as a rigid-leg connected to a universal joint the enables motion along two rotational DOFs (lift and swing); we call this model \hamrsimple (\secref{sec:hamr-simple}, \figref{fig:trans_schematic}). This model, however, is a poor approximation of the SFB-transmission's dynamics, and consequently, policies generated using this model are unlikely to transfer to the physical robot.

\subsection{\hamrfull transmission model}
\label{sec:hamr-full}

We model the SFB transmission using the pseudo-rigid body approximation \cite{howell2001compliant}, with the flexures and carbon fiber linkages modeled as pin joints and rigid bodies, respectively. Each flexure is assumed to deflect only in pure bending with its mechanical properties described by a torsional spring and damper that are sized according to the procedure described in \cite{doshi2015model}. Given these assumptions, an intuitive dynamical description of each SFB transmission has two inputs (voltages applied to the actuators), eight generalized coordinates (two independent coordinates and six dependent), and six constraints imposed by the kinematics of the three parallel chains. Mathematically, we can write this as:
\begin{align}
\label{eqn:forced_lagrangian}
\mathbf{M}(\mathbf{q}) \ddot{\mathbf{q}} + \mathbf{C}(\mathbf{q}, \dot{\mathbf{q}}) \dot{\mathbf{q}} + \mathbf{g}(\mathbf{q}) + 
\mathbf{H}(\mathbf{q})^T \boldsymbol{\lambda} & =  \mathbf{f}^\textrm{act}(\mathbf{q}, \mathbf{u}) & 
\\
\label{eqn:loop_const}
\mathbf{h}(\mathbf{q})& =  \mathbf{0}.
\end{align}
Here $\mathbf{q} \in \mathbb{R}^{8}$ is the vector of generalized coordinates, consisting of the actuator and flexure deflections, $\boldsymbol{\lambda} \in \mathbb{R}^{6}$ are the constraint forces, $\mathbf{u} \in \mathbb{R}^2$ is the control input (actuator voltage), and the dot superscripts represent time derivatives.  Moreover, $\mathbf{M}$, $\mathbf{C}$, and $\mathbf{g}$ are the inertial, Coriolis, and potential terms in the manipulator equations,  $\mathbf{H}(\mathbf{q})^T = (\partial \mathbf{c} / \partial \mathbf{q})^T$ is the Jacobian mapping constraint forces into generalized coordinates, and $\mathbf{f}^{\textrm{act}}$ is the vector of generalized actuator forces. Finally, equation \eqref{eqn:loop_const} enforces the kinematic-loop constraints, $\mathbf{h}(\mathbf{q})$. Together, \eqref{eqn:forced_lagrangian} and \eqref{eqn:loop_const} form a system of differential-algebraic equations whose solution can be expensive to compute because of the non-linear kinematic constraints and the added (dependent) generalized coordinates. In the future sections, we will refer to this model as $\mathbf{\ddot{q}} = \mathbf{f}(\mathbf{q}, \mathbf{\dot{q}}, \mathbf{u})$.

\subsection{\hamrsimple transmission model}
\label{sec:hamr-simple}
A \hamrsimple transmission is modeled using rigid-body dynamics. Unlike the SFB-transmission, this model only has two DOFs (lift and swing angles), and is driven by two-inputs (torques directly applied to the legs). The model for a \hamrsimple transmission can be written as 
\begin{align}
\label{eqn:hamr_simple}
\mathbf{M_s}(\mathbf{q_s}) \ddot{\mathbf{q_s}} + \mathbf{C_s}({\mathbf{q_s}}, \dot{{\mathbf{q_s}}}) \dot{{\mathbf{q_s}}} +  \mathbf{{g_s}}(\mathbf{{q_s}}) 
 =  {\mathbf{u_s}}.
\end{align}
Here ${\mathbf{q_s}} \in \mathbb{R}^{2}$ is the vector of generalized coordinates, consisting of $\theta_\text{l}$ and $\theta_\text{s}$, and ${\mathbf{u_s}} \in \mathbb{R}^2$ is the control inputs (hip torques). As before, ${\mathbf{M_s}}$, ${\mathbf{C_s}}$, and ${\mathbf{g_s}}$ are the inertial, Coriolis, and potential terms in the manipulator equations. In the future sections, we will refer to this model as $\mathbf{\ddot{q}_s} = \mathbf{f_s}(\mathbf{q_s}, \mathbf{\dot{q}_s}, \mathbf{u_s})$. These equations are less expensive to evaluate as they are both lower dimensional and unconstrained. 

\subsection{Leg Control}
\label{subsec:leg_control}

We run a leg position controller on top of both robot models, producing actuator voltage updates at \SI{20}{\kilo\hertz} and accepting target joint angles ($\mathbf{q} \in \mathbb R^8$) at \SI{250}{\hertz}.  Evaluating learned policies at high frequencies is computationally-intensive and becomes especially infeasible in embedded contexts; joint-space control allows stable low-frequency input schemes while attempting to sample torque or voltage targets at lower frequencies often leads to poor mobility or unstable behaviors.  

\section{Method}
\label{sec:method}
We propose a data-driven modeling methodology \textit{residual model learning} to build a platform for learning microrobot locomotion behaviors. Conceptually, we recognize that traditional analytic rigid-body models succeed at approximately computing motion and collision between rigid bodies. Our approach builds off of this base as a prior and allows us to augment our models when rigid-body assumptions no longer hold.  We do this by learning a model that minimizes the residual in dynamics between \hamrfull and \hamrsimple, using \hamrsimple as our rigid body prior.  This hybrid approach is more sample efficient and grants us additional robustness for the eventual transfer of learned policies back to \hamrfull.

\subsection{Dynamics modeling via Residual Learning}
\label{sec:residual-model}

The goal of dynamics modeling is to produce a function $\mathbf{f}$ that, given the current system state $\mathbf{x}=[\mathbf{q},\dot{\mathbf{q}} ]$ and a control input $\mathbf{u}$, accurately predicts accelerations $\ddot{\mathbf{q}}$.  Our method takes advantage of \hamrsimple by computing its dynamics $\mathbf{f}_\text{s}(\mathbf{q}_s, \dot{\mathbf{q}}_s, \mathbf{u}_s)$ over \hamrsimple state $\mathbf{q_s}$ and learning a feedback function $\mathbf{\hat{k}}$ over the \hamrfull input $\mathbf{u}$ such that:

\begin{equation}
\label{eq:learned_hamrfull}
\vspace{-0.1cm}
\mathbf{f}_\text{s}\big(\mathbf{q_\text{s}}, \dot{\mathbf{q}}_\text{s},  \mathbf{\hat{k}}(\mathbf{q_\text{s}}, \dot{\mathbf{q}}_\text{s}, \mathbf{u}) \big) \approx \mathbf{f}(\mathbf{q}, \dot{\mathbf{q}}, \mathbf{u}).
\end{equation}

We know such a function $\mathbf{\hat{k}}$ exists, as our hip sub-states are fully actuated. Thus our dynamics $\mathbf{f}$ must be surjective, so there always exists some $\mathbf{u}$ that produces the desired response. To learn our particular $\mathbf{\hat k}$, we seek to minimize the $l^2$-norm residuals between $\mathbf{f}_\text{s}$ and the true dynamics $\mathbf{f}$ over white noise voltage inputs $\mathbf{u}$:

\begin{align}
\label{eqn:optim_k}
\mathbf{\hat k} = \argmin_{\mathbf{\hat k}} ||&\mathbf{f_\text{s}}(\mathbf{q}_s, \dot{\mathbf{q}}_s, \mathbf{\hat k}(\mathbf{q}_s, \dot{\mathbf{q}}_s, \mathbf{u})) - \mathbf{f}(\mathbf{q}, \dot{\mathbf{q}}, \mathbf{u})||_2^2.
\end{align}

We use Adam~\cite{kingma2014adam}, a variant of gradient descent optimizer, to optimize Equation \eqref{eqn:optim_k}. As the simulator itself is not differentiable, we use a neural network $\mathbf{\hat f_\text{s}}$ to approximate the simplified dynamics model $\mathbf{f_\text{s}}$ in order to optimize $\mathbf{\hat k}$. In our experiments, both $\mathbf{\hat f_\text{s}}$ and $\mathbf{\hat k}$ are parameterized by a fully-connected neural network with 2 hidden layers of 100 neurons each and $\tanh$ activations, due to all variables being normally distributed and scaled to fit in the interval $[-1,1]$.  Additional network architectures were not explored, as this architecture was found to produce a sufficient quality of results.

\subsection{Multi-step recursive training process}

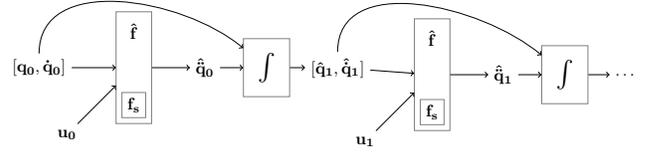
\begin{figure}[t]
\centering
\scalebox{0.63}{
\begin{tikzpicture}
  \node (f_0_l) { $\mathbf{\hat{f}}$ };
  \node (s_0) [below=of f_0_l, rectangle, draw=black!50, ] { $\mathbf{f_{s}}$};
  \node (f_0) [draw=black!50, fit={(s_0) (f_0_l)}] {};
  
  \node (q_0) [left of=f_0, node distance=20mm] {$[\mathbf{q_0}, \mathbf{\dot q_0}]$};
  \node (u_0) [below left of=f_0, node distance=20mm] {$\mathbf{u_0}$};
  \draw[->] (q_0) -- (f_0) ;
  \draw[->] (u_0) -- (f_0) ;
  
  \node (qdot_0) [right of=f_0, node distance=15mm] {$\mathbf{\hat{\ddot q}_0}$};
  \draw[->] (f_0) -- (qdot_0) ;
  
  \node (int_0) [rectangle, right of=qdot_0, node distance=13mm, draw=black!50, inner sep=3mm] {\LARGE $\int$};
  \draw[->] (qdot_0) -- (int_0);
  \draw[->] (q_0)  to [out=90,in=140] (int_0);
  
  \node (q_1) [right of=int_0, node distance=15mm] {$[\mathbf{\hat{q}_1}, \mathbf{\hat{\dot{q}}_1}]$};
  \draw[->] (int_0) -- (q_1);

  \node (f_1_l) [above right=0mm and 11mm of q_1] { $\mathbf{\hat{f}}$ };
  \node (s_1) [below=of f_1_l, rectangle, draw=black!50, ] { $\mathbf{f_{s}}$ };
  \node (f_1) [draw=black!50, fit={(s_1) (f_1_l)}] {};
  \node (u_1) [below left of=f_1, node distance=20mm] {$\mathbf{u_1}$};
  \draw[->] (q_1) -- (f_1) ;
  \draw[->] (u_1) -- (f_1) ;

  \node (qdot_1) [right of=f_1, node distance=15mm] {$\mathbf{\hat{\ddot q}_1}$};
  \draw[->] (f_1) -- (qdot_1) ;
  
  \node (int_1) [rectangle, right of=qdot_1, node distance=13mm, draw=black!50, inner sep=3mm] {\LARGE $\int$};
  \draw[->] (qdot_1) -- (int_1);
  \draw[->] (q_1)  to [out=90,in=140] (int_1);
  
  \node (etc) [right of=int_1, node distance=13mm] {$\dots$};
  \draw[->] (int_1) -- (etc);

\end{tikzpicture}
}
\caption{Block diagram of multi-step Euler-integrated model execution.  Given a series of actuator voltage inputs $\mathbf{u_{0\dots t}}$ and an initial robot position $\mathbf{q_0}$ we produce trajectories $\mathbf{\hat q_{0\dots t}}$, $\mathbf{\hat {\dot q}_{0\dots t}}$ and $\mathbf{\hat{\ddot q}_{0\dots t}}$ (positions, velocities, and accelerations) that we then supervise from data.}
\label{fig:multi-step}
\vspace{-0.2cm}
\end{figure}

To overcome compounding tracking errors when optimizing over single-step horizons, we modify training to predict HAMR's trajectory for a time horizon of $h (>1)$ time steps. In order to make predictions over $h$ steps, we integrate $\hat{\dot{\mathbf{x}}}_t = [\hat{\dot{\mathbf{q}}}_t, \hat{\ddot{\mathbf{q_t}}}]$ at each time step $t$ to produce a $\hat{\mathbf{x}}_{t+1}$ that is fed back into the model along with $\mathbf{u}_{t+1}$ to produce HAMR's state at the next time step (\figref{fig:multi-step}). This generates a trajectory: \{$\hat{\mathbf{q}}_{t:t+h},\hat{\dot {\mathbf{q}}}_{t:t+h},\hat{\ddot{\mathbf{q}}}_{t:t+h}$\}, that is fit against the ground truth. In practice, we train over a horizon of $h=\SI{8}{\milli\second}$ using a batch size of $8$ and Adam optimizer \cite{kingma2014adam}, as this was found to produce a high quality of fit.

\subsection{Modeling Results}

\begin{figure}[t]
    \centering
    \includegraphics[width=0.95\columnwidth]{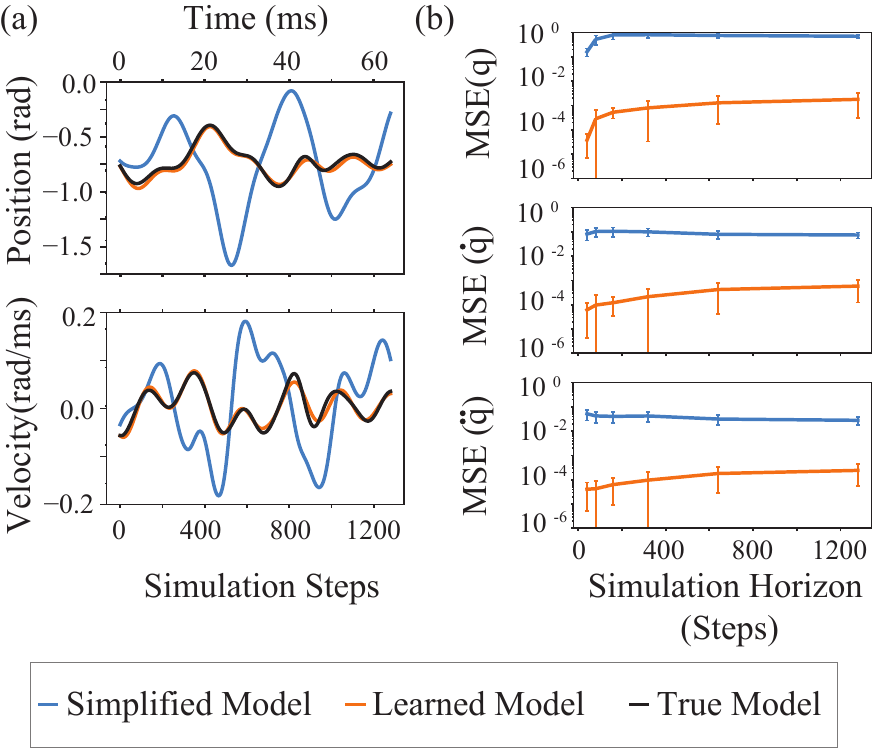}
    \caption{Evaluation of the learned (residual) model. (a) Example time-traces of position (top) and velocity (bottom). The learned model (orange) closely matches the true \hamrfull model (black). (b) Mean-squared validation error (MSE) in position (top), velocity (middle), and acceleration (bottom) as a function of the simulation horizon. The learned model (orange) exhibits significantly lower error than the simplified \hamrsimple model (blue).} 
    \label{fig:res_val}
    \vspace{-0.2cm}
\end{figure}

Our hybrid modeling methodology achieves extremely low error in tracking \hamrfull trajectories over a range of simulation horizons using only 12 seconds of passively collected white noise input trajectories. Qualitatively, our model achieves nearly perfect tracking to previously unseen \hamrfull trajectories (\figref{fig:res_val}a). Quantitatively, our method achieves orders of magnitude lower mean-squared error (MSE) than \hamrsimple (\figref{fig:res_val}b).  Our model learning methodology is entirely agnostic to different contact surfaces and environments, as our hybrid approach allows contacts to be added to the model at a later stage.

\subsection{Learning Primitives}

\begin{figure*}[t!]
    \centering
    \includegraphics[width=0.92\textwidth]{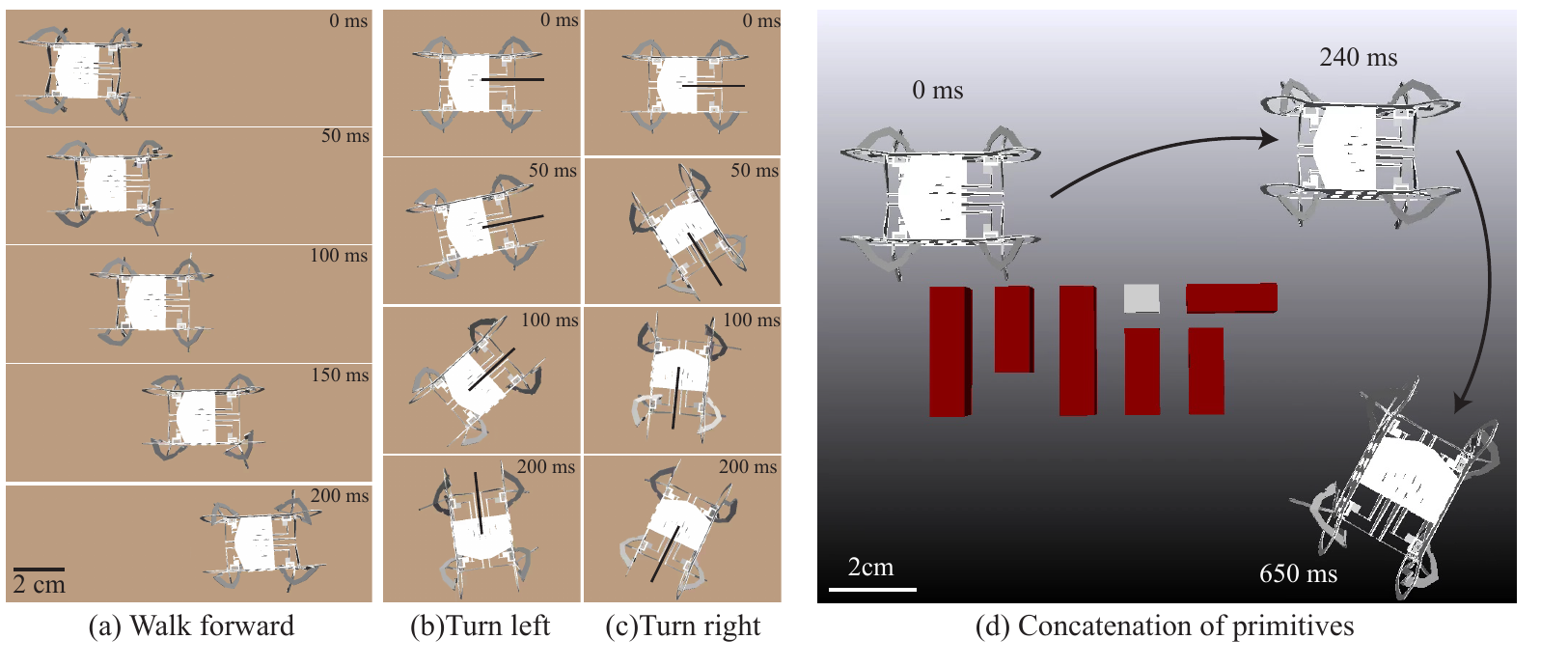}
    \caption{Roll-outs of policies learned using our residual modeling approach evaluated on the true \hamrfull. (a) Walking and (b, c) turning policies produce realistic behaviors that experience minimal loss in agility through transfer. (d) We manually concatenate policies to achieve more complex locomotion: \hamrfull moves around an obstacle by walking forward, turning right, and then continuing to walk forward.}
    \label{fig:rollout}
    \vspace{-0.2cm}
\end{figure*}

To evaluate our modeling methods' robustness and accuracy, we train locomotion primitives for walking and turning in two directions using Proximal Policy Optimization (PPO) \cite{schulman2017proximal}, a general model-free reinforcement learning algorithm for continuous control tasks.  While model-free deep reinforcement learning algorithms like PPO are powerful enough to discover highly dynamic locomotion behaviors \cite{tan2018sim}, they are also vulnerable to over-fitting to inaccurate or exploitable simulation environments \cite{amodei2016concrete}.

Our parametric walking primitive is defined by the following reward function:

\begin{equation}
r_{walking} = 2p_d \dot x - \sqrt{ \theta_\text{roll}^2  + \theta_\text{pitch}^2 },
\end{equation}
where $p_d \in \{-1, 1\}$ is the commanded walking direction, $\dot x$ is the robot body velocity in the $x$-direction, and the latter term is a stabilizing expression to encourage upright policies.  Our turning primitive is defined by the following reward function:
\begin{equation}
r_{turning} = 0.5 p_d \dot \theta_\text{yaw},
\end{equation}
where $p_d$ is defined above and $ \dot \theta_\text{yaw}$ is the robot body angular velocity in the yaw axis. Roll-outs of each primitive is ended after \SI{300}{\milli\second} of simulation time has elapsed, or a simulator convergence error has occurred (due to the robot reaching an unstable state).  Primitive policies accept as input the robot state, consisting of the body position, leg angles, body velocity, and leg angular velocities, and produce angle targets for the joint-space controller (see \secref{subsec:leg_control}).  These policies are queried at \SI{250}{\hertz}.

\section{Control Experiments and Results}
\label{sec:experiments}
We evaluate our residual learned \hamrfull model by (a) using it as a simulation environment to train parametric locomotion primitives using PPO, (b) evaluating learned rewards, and (c) observing quality of transfer when policies are then evaluated on the true robot model. We treat the true robot model as a proxy for the real world, as we hope to repeat these experiments using real robot data and real robot evaluation rather than \hamrfull.  Successful transfer of PPO policies indicates accurate and robust dynamics learning due to the powerful capacity of model-free methods to maximize task rewards, often circumventing environment or reward design intention to do so. For links to videos of learned HAMR locomotion behaviors, see our project website: \url{https://sites.google.com/view/hamr-icra-2021}

\begin{figure*}[t]
    \centering
    \includegraphics[width=0.88\textwidth]{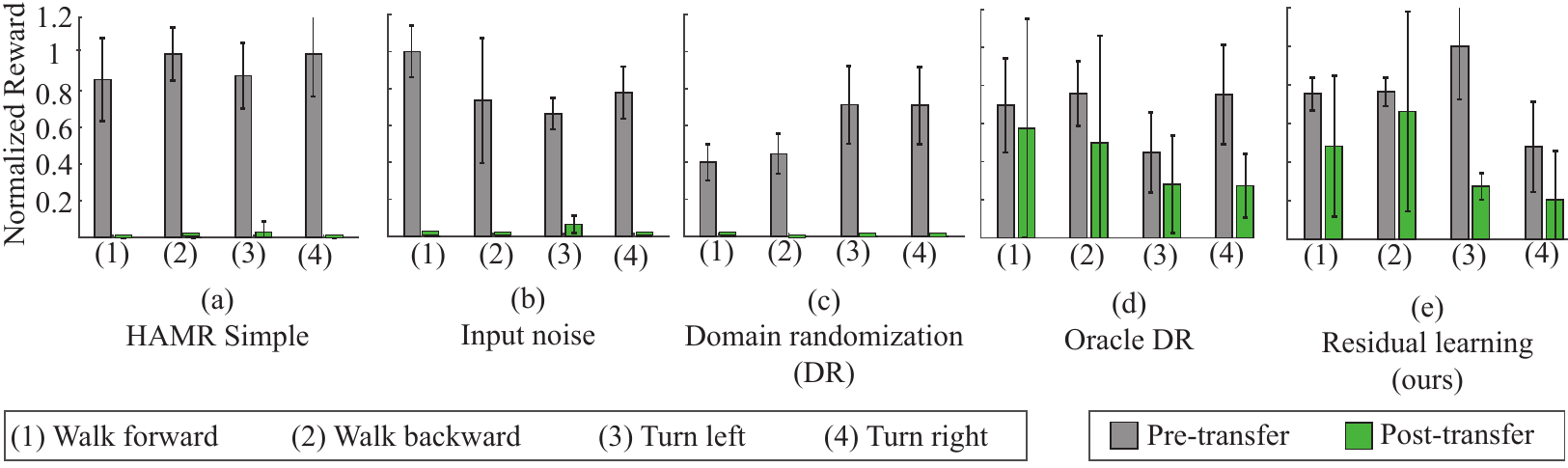}
    \caption{Policy rewards for walking and turning pre-transfer on the training environment (green) and post-transfer on \hamrfull (grey).  Our method achieves less degradation post-transfer than other methods not privy to privileged knowledge, far outperforming our \hamrsimple baseline, adding noise to inputs, and domain randomization.  Rewards are normalized to the maximum achieved for each primitive and are clipped at zero reward. Standard deviations are computed across 25 trials (5 random seeds with 5 roll-outs per seed).}
    \label{fig:barchart}
    \vspace{-0.3cm}
\end{figure*}

\subsection{Comparisons}

We compare our method against five other approaches, representing the common modeling methods for learning robust transferable locomotion controllers. We note that it is not feasible to compare against policies trained using \hamrfull, as each experiment would take $\approx 25$ days using a 48-core machine due to the model's complex kinematic constraints.

\textbf{\hamrsimple}: As a baseline, we evaluate transfer from a controller learned on the lumped-parameter simulation environment prescribing PD controller targets.

\textbf{End-to-end modeling}: We evaluate learning an end-to-end \hamrfull model with contact using a similar methodology to other model-based solutions \cite{nagabandi2018neural}.  We use the same two-layer 100-neuron MLP as we use in our residual modeling and learn to predict state differences between timesteps, using 10$\times$ the data provided to RML.

\textbf{Noisy Inputs}: One simple method for increasing the robustness of learned controllers is to add noise to simulator inputs \cite{tan2018sim}.  We evaluate this approach by adding Gaussian noise of $\mu=0$, $\sigma=0.01$ to \hamrsimple torque commands produced by the PD controller, with parameters chosen to maximize transfer performance.

\textbf{Domain Randomization}: The key difference between \hamrsimple and \hamrfull is that the former doesn't correctly model the state-dependent variation in the inertial, Coriolis, and potential terms on the LHS of the \eqref{eqn:forced_lagrangian}. These un-modeled effects can be primarily attributed to configuration-dependent changes in the SFB-transmission's inertia tensor and generalized stiffness. Here we randomly vary the inertia tensor and generalized stiffness of \hamrsimple between trials by adding multiplicative Gaussian noise of $\mu=1$, $\sigma=0.2$ clipped at $[0.7,1.3]$ to the inertia matrix and stiffnesses.  We ensure that the perturbed inertia matrix is positive definite and satisfies the triangle inequality for physical plausibility.

\textbf{Oracle Domain Randomization}: For this method we sample directly from the distribution of \hamrfull's inertia tensors and generalized stiffness (measured during hard-coded locomotion behaviors). The lumped inertia of \hamrfull's transmission is calculated as a function of its configuration using the parallel axis theorem. The generalized stiffness is projected from \hamrfull's generalized coordinates onto \hamrsimple's generalized coordinates by using the Jacobian that maps between the two.  We call this method ``oracle'' domain randomization as it uses privileged information from \hamrfull that would be inaccessible on the real robot. 

\textbf{Model-Based RL (PETS)}: Finally, we compare the compare data-efficiency of learning residual model v/s learning policies on HAMR-full via a state-of-the-art model-based reinforcement learning algorithm, PETS~\cite{chua2018deep}. Because this method is too slow to train using \hamrfull, we instead use our learned \hamrfull as the proxy simulator (see LHS of Equation \eqref{eq:learned_hamrfull}). Since we are primarily interested in data-efficiency comparison, the choice of learned \hamrfull instead of actual \hamrfull does not effect the conclusions.  

\vspace{-0.1cm}
\subsection{Results}
Our approach produces policies with learned locomotion behavior preserved through transfer to \hamrfull (\figref{fig:rollout}). In particular, we learn four different primitives: walking forward and backward as well as turning left and right. We find that the average speed during forward and backward locomotion is 4.62 $\pm$ 1.60 body-lengths per second and 3.18 $\pm$ 1.62 body-lengths per second, respectively. Similarly, average turning speed during right and left turns is 628.26 $\pm$ 193.99 \textdegree/s and 536.86 $\pm$ 201.85 \textdegree/s, respectively. We note that these speeds are consistent with those previously observed during locomotion with the physical robot~\cite{goldberg2017highb}. 

We also evaluate our approach against the five baselines, and find that our approach achieves markedly better reward when evaluated on \hamrfull (\figref{fig:barchart}).
The \hamrsimple baseline (\figref{fig:barchart}a) performs poorly, achieving low or negative reward upon transfer. This verifies our hypothesis that, even with a low-level controller, it is difficult to transfer policies trained on the simplified model directly. Similarly, adding input noise (\figref{fig:barchart}b) and domain randomization  (\figref{fig:barchart}c) also performs poorly after the transfer, achieving close to zero or negative reward. The poor performance of these randomization methods indicates that it might be difficult to achieve good transfer results on microrobots using naive (uniform or Gaussian) domain randomization.   

Oracle Domain Randomization  (\figref{fig:barchart}d) uses privileged information about the dynamics of \hamrfull and performs similarly to our method. However, such information cannot be obtained for actual robots. This further suggests that it is necessary to develop approaches that are able to leverage simulators for data efficiency but go beyond domain randomization: ours is one such method. 

We did not plot results of the baseline that learned the dynamics end-to-end to create a proxy simulator that was then used to train PPO. This was because, the learned model was inaccurate and was therefore exploited by PPO. While, the pre-transfer rewards were orders of magnitude greater than other approaches these policies failed to transfer and achieved near-zero reward post-transfer. In contrast, our method can learn an accurate dynamics model using offline data by simply modelling the residual. 

We also find that when compared to PETS, our method requires 10x less interaction data.  PETS had much higher variance (more than twice of our method) and on average achieved 0.9x the maximum reward (post-transfer) achieved our method; see appendix for detailed results.
\section{Conclusion}

We present a hybrid modeling approach for the HAMR  capable of leveraging a simple yet inaccurate rigid-body model along with a learned corrective component to produce a fast and accurate simulation environment for reinforcement learning. Our model achieves tight trajectory tracking, and is robust enough to be used as a simulation environment to train locomotion policies that transfer to our more complex model with no fine-tuning. Our method outperforms common strategies used to build robust transferable controllers, such as adding model noise and performing domain randomization. We believe this approach of combining simple rigid-body models with learned components is a promising method for modeling microrobots, due to their use of non-rigid materials in complex linkage systems. As a next step, we plan to train our residual model using data from a physical HAMR and attempt to transfer our learned policies onto the physical robot. If successful, we can then use residual model learning as a method for efficiently modeling the transmission mechanisms used in other millimeter-scale robots and devices.

\section{Acknowledgements}
This work as in part supported by the DARPA Machine Common Sense Grant and the MIT-IBM Quest program. We thank the MIT SuperCloud and Lincoln Laboratory Supercomputing Center for providing (HPC, database, consultation) resources that have contributed to the research results reported within this paper/report.

\appendix

\section{Model-based Baseline}

\begin{figure}[ht]
    \centering
    \includegraphics[width=0.95\columnwidth]{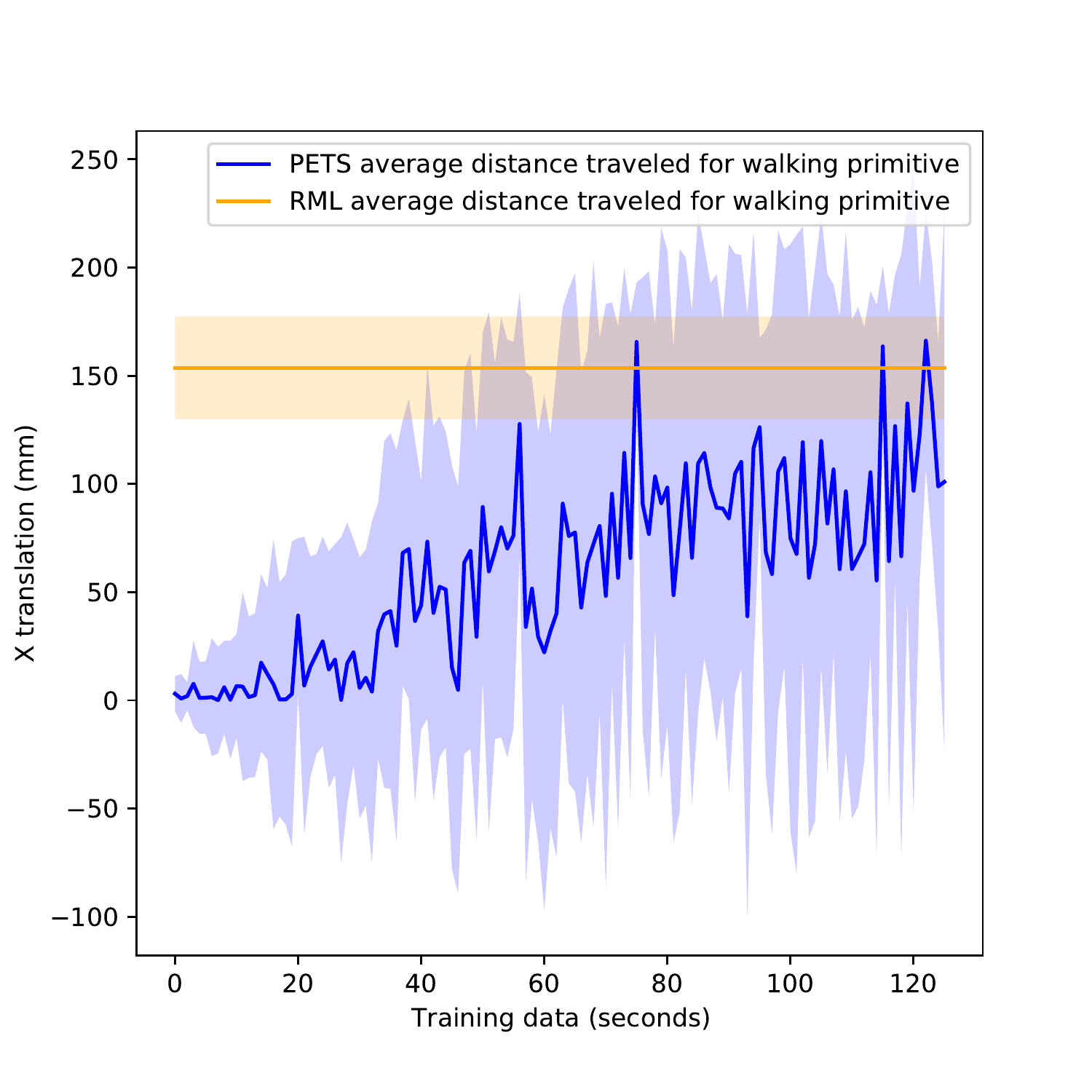}
    \caption{Learning curve of PETS learning a bi-directional walking primitive on our residual model (as a proxy for \hamrfull).  PETS is unable to robustly learn the bidirectional nature of the policy (demonstrating high variance), takes ~10x as much data as our method, and only achieves $\sim$ 90\% of the translational movement over each episode after convergence.} 
    \label{fig:pets-mean-reward}
    \vspace{-0.2cm}
\end{figure}

In order to compare our method against model-based approaches, we evaluate PETS~\cite{chua2018deep} on our hybrid simulator to produce an approximate measure of sample efficiency (\figref{fig:pets-mean-reward}), as evaluating on \hamrfull would be impractical.  Note that data collected through PETS is on-policy, while our method uses offline data that is not task specific.

\bibliography{ref, references, agile}

\end{document}